\documentclass[10pt,twocolumn,letterpaper]{article}

\usepackage[pagenumbers]{cvpr} 

\usepackage[utf8]{inputenc} 
\usepackage[T1]{fontenc}    
\usepackage{url}            
\usepackage{booktabs}       
\usepackage{amsfonts}       
\usepackage{nicefrac}       
\usepackage{microtype}      
\usepackage{xcolor}         
\usepackage{comment}

\usepackage{graphicx}
\usepackage{booktabs}

\usepackage[accsupp]{axessibility}  

\usepackage{wrapfig,lipsum}
\usepackage[utf8]{inputenc} 
\usepackage[T1]{fontenc}    
\usepackage{url}            
\usepackage{booktabs}       
\usepackage{amsfonts}       
\usepackage{nicefrac}       
\usepackage{microtype}      
\usepackage{xcolor,colortbl}         
\usepackage{multirow} 
\usepackage{subcaption}
\usepackage{times}
\usepackage{epsfig}
\usepackage{graphicx}
\usepackage{caption}
\usepackage{amsmath}
\usepackage{amssymb}
\usepackage{booktabs}
\usepackage{wrapfig}
\usepackage{array}
\usepackage{algorithmic}
\usepackage[ruled]{algorithm2e}

\usepackage[accsupp]{axessibility}  

\newcolumntype{P}[1]{>{\centering\arraybackslash}p{#1}}
\definecolor{cvprblue}{rgb}{0.21,0.49,0.74}
\usepackage[pagebackref,breaklinks,colorlinks,allcolors=cvprblue]{hyperref}

\def\paperID{17000} 
\def\confName{CVPR}
\def\confYear{2025}

\title{Soft Self-labeling and Potts Relaxations \\ for Weakly-Supervised Segmentation}

\author{ Zhongwen (Rex) Zhang \hspace{1ex} \& 
\hspace{1ex} Yuri Boykov \\
University of Waterloo, Canada \\
{\tt\small z889zhan@uwaterloo.ca, yboykov@uwaterloo.ca}
}

\begin{document}
\def\BL{\text{\tiny\sc BL}}
\def\Q{\text{\tiny\sc Q}}
\def\NQ{\text{\tiny\sc NQ}}
\def\CCE{\text{\tiny\sc CCE}}
\def\LQ{\text{\tiny\sc LQ}}
\def\CD{\text{\tiny\sc CD}}
\def\CE{\text{\tiny\sc CE}}
\def\RCE{\text{\tiny\sc RCE}}

\maketitle

\begin{abstract}
  We consider weakly supervised segmentation where only a fraction of pixels have ground truth labels (scribbles) and focus on a self-labeling approach optimizing relaxations of the standard unsupervised CRF/Potts loss on unlabeled pixels. While WSSS methods can directly optimize such losses via gradient descent, prior work suggests that higher-order optimization can improve network training by introducing hidden pseudo-labels
  and powerful CRF sub-problem solvers, e.g. graph cut. However, previously used hard pseudo-labels can not represent class uncertainty or errors, which motivates soft self-labeling. We derive a principled auxiliary loss and systematically evaluate standard and new CRF relaxations (convex and non-convex), neighborhood systems, 
  and terms connecting network predictions with soft pseudo-labels. 
  We also propose a general continuous sub-problem solver. Using only standard architectures, soft self-labeling consistently improves scribble-based training and outperforms significantly more complex specialized WSSS systems. It can outperform full pixel-precise supervision. 
  Our general ideas apply to other weakly-supervised problems/systems. Code can be found on \url{https://vision.cs.uwaterloo.ca/code}. \\[0.5ex]
  {\em\footnotesize {\bf keywords}: soft pseudo-labels, collision cross-entropy, collision divergence}
\end{abstract}\\[-7.5ex]

\section{Introduction}
\label{sec: introduction}

Full supervision for semantic segmentation requires thousands of training images with complete pixel-accurate ground
truth masks. Their  high costs explain the interest in weakly-supervised approaches based on image-level
class {\em tags} \cite{kolesnikov2016seed, araslanov2020single}, pixel-level {\em scribbles}
\cite{lin2016scribblesup,tang2018regularized,tang2018normalized}, or {\em boxes} \cite{Kulharia2020Box2SegAW}. 
This paper is focused on weak supervision with {\em scribbles}, which we also call {\em seeds} or {\em partial masks}. While only slightly more expensive than image-level class tags, scribbles on less than $3\%$ of 
pixels were previously shown to achieve accuracy approaching full supervision without any modifications of the segmentation models. In contrast, tag supervision typically requires highly specialized systems and complex multi-stage training procedures, which are hard to reproduce.
Our interest in the scribble-based approach is motivated by its practical simplicity and 
mathematical clarity. The corresponding methodologies are focused on the design of 
unsupervised or self-supervised loss functions and stronger optimization algorithms. 
The corresponding solutions are often general and can be used in different weakly-supervised applications.

\subsection{Scribble-supervised segmentation}

Let $\Omega$ denote a set of image pixels and $S\subset\Omega$ denote a subset of pixels with ground truth labels, 
i.e. {\em seeds} or {\em scribbles} typically marked by mouse-controlled
UI for image annotations. One example of image scribbles is in Fig.\ref{Fig: pairwise illustration}(a). 
The ground truth label at any given pixel $i\in S$ is an integer 
\begin{equation} \label{eq:GT_index}
\bar{y}_i\in\{1,\dots,K\}
\end{equation}
where $K$ is the number of classes including the background. Without much ambiguity, 
it is convenient to use the same notation $\bar{y}_i$ for the equivalent {\em one-hot} distribution 
\begin{equation} \label{eq:GT_onehot}\resizebox{0.43\textwidth}{!}{$
\bar{y}_i \;\;\equiv\;\;(\bar{y}_i^1,\dots,\bar{y}_i^K)\,\in\Delta^K_{0,1} \quad 
\text{for}\quad \bar{y}_i^k := \left[k=\bar{y}_i\right]$}
\end{equation}
where $[\,\cdot\,]\in\{0,1\}$ is the {\em True} operator.  
Set $\Delta^K_{0,1}$ represents $K$ possible one-hot distributions, which are vertices of the $K$-class 
{\em probability simplex} $$\Delta^K\;\;:=\;\;\{p=(p^1,\dots,p^K)\;|\;p^k\geq 0,\;\sum_{k=1}^K p^k = 1\}$$ 
representing all $K$-categorical distributions. The context of each specific expression 
should make it obvious if $\bar{y}_i$ is a class index \eqref{eq:GT_index} or 
the corresponding one-hot distribution \eqref{eq:GT_onehot}.

Loss functions for scribble-supervised segmentation typically use {\em negative log-likelihoods} (NLL) over scribbles $i\in S\subset\Omega$ with ground truth labels $\bar{y}_i$
\begin{equation} \label{eq:NLL}
-\sum_{i\in S} \ln\sigma_i^{\bar{y}_i} 
\end{equation}
where $\sigma_i=(\sigma_i^1,\dots,\sigma_i^K)\in\Delta^K$ are model predictions at pixels $i$. 
This loss is also standard in full supervision where $S=\Omega$. In this case, no other losses are necessary for training. However, in a weakly supervised setting the majority of pixels are unlabeled, and unsupervised losses 
are needed for $i\not\in S$. 

The most common unsupervised loss in image segmentation is the Potts model and its relaxations. It is a pairwise loss defined on pairs of {\em neighboring} pixels $\{i,j\}\in{\cal N}$ 
for a given neighborhood system ${\cal N}\subset\Omega\times\Omega$, typically corresponding to the 
{\em nearest-neighbor} grid (NN) \cite{boykov2001interactive,grady2006random}, or other {\em sparse} (SN) \cite{veksler2022sparse} or
{\em dense} neighborhoods (DN) \cite{krahenbuhl2011efficient}. The original Potts model is defined for discrete segmentation
variables, e.g. as in
$$\sum_{\{i,j\}\in{\cal N}} P(\sigma_i,\sigma_j)\quad\quad\text{where}\quad P(\sigma_i,\sigma_j) = [\sigma_i\neq\sigma_j] $$
assuming integer-valued one-hot predictions $\sigma_i\in\Delta^K_{0,1}$. This {\em regularization} loss 
encourages smoothness between the pixels. Its popular {\em self-supervised} variant is 
$$  P(\sigma_i,\sigma_j) = w_{i,j}\cdot [\sigma_i\neq\sigma_j] $$
where pairwise affinities $w_{ij}$ are based on local intensity edges \cite{boykov2001interactive,grady2006random,krahenbuhl2011efficient}.
Of course, in the context of network training, one should use relaxations of $P$ applicable to 
soft predictions $\sigma_i\in\Delta^K$. Many forms of such relaxations \cite{Ravikumar:ICML2006,Zach:CVPR2012} 
were studied in segmentation, e.g. {\em quadratic} \cite{grady2006random}, {\em bi-linear} \cite{tang2018regularized}, {\em total variation} \cite{Pock:CVPR2009,chambolle2011first}, and others \cite{couprie2010power}. 

Another general unsupervised loss relevant to training segmentation networks is the entropy of predictions, which is also known as {\em decisiveness} \cite{bridle1991unsupervised, grandvalet2004semi} 
$$ \sum_i H(\sigma_i)$$ where $H$ is the Shannon's entropy function. 
This loss is known to improve generalization and the quality of representation by moving (deep) features 
away from the decision boundaries \cite{grandvalet2004semi}. Widely known in the context of unsupervised or semi-supervised classification, this loss also matters in weakly-supervised segmentation where it is used 
explicitly or implicitly\footnote{A unary decisiveness-like term is the difference 
between quadratic and bi-linear ({\em tight} but non-convex) relaxations 
\cite{Ravikumar:ICML2006,marin2021robust} of the Potts model.}\@.

Weakly-supervised segmentation methods \cite{wang2019boundary, pan2021scribble, ke2021universal, chen2021seminar} 
also use other unsupervised losses (e.g. contrastive), clustering criteria (e.g. K-means), highly-specialized architectures, and ad-hoc training procedures.
Our goal is to demonstrate that better results can be easily achieved on standard architectures
by combining the standard losses above
\begin{equation} \label{eq:loss_ws}\resizebox{0.43\textwidth}{!}{$
-\sum_{i\in S} \ln\sigma_i^{\bar{y}_i}\;+\;\eta\sum_{i\not\in S} H(\sigma_i)
\;+\;\lambda\sum_{ij\in {\cal N}} P(\sigma_i,\sigma_j)$}.
\end{equation}
but replacing basic gradient descent optimization \cite{tang2018regularized,veksler2022sparse}
by higher-order {\em self-labeling} techniques \cite{lin2016scribblesup,marin2019beyond,marin2021robust}
incorporating optimization of auxiliary {\em pseudo-labels} as sub-problems.

\subsection{Self-labeling and hard pseudo-labels} 

One motivation for the self-labeling approach to weakly-supervised segmentation comes from 
well-known limitations of gradient descent for optimizing the Potts relaxations.
Even convex relaxations \cite{grady2006random,Pock:CVPR2009,chambolle2011first} could be
challenging in combination with the concave entropy term in \eqref{eq:loss_ws}.

Typical self-labeling methods, including one of the first works on scribble-based semantic segmentation
\cite{lin2016scribblesup}, introduce a sub-problem focused on the estimation of {\em pseudo-labels} 
for unlabeled points. This subproblem can be formally ``derived'' from the original network optimization problem or 
``designed'' as a heuristic, see the overview below. 
We focus on principled auxiliary loss formulations where network parameters and pseudo-labels 
are jointly optimized in a provably convergent manner. In this case pseudo-labels work as hidden optimization variables
simplifying the model estimation, e.g. as in the EM algorithm for GMM. 

We denote pseudo-labels $y_i$ differently from the ground truth labels $\bar{y}_i$
by omitting the bar. It is important to distinguish them since $\bar{y}_i$ for $i\in S$
are given, while $y_i$ for $i\in \Omega\backslash S$ are estimated.
The majority of existing self-labeling methods \cite{lin2016scribblesup,ahn2018learning,marin2019beyond,ahn2019weakly,lee2019ficklenet,marin2021robust,wang2023treating} 
estimate {\em hard} pseudo-labels, which could be equivalently 
represented either by class indices
\begin{equation} \label{eq:label_index}
y_i\in\{1,\dots,K\}
\end{equation}
or by the corresponding one-hot categorical distributions
\begin{equation} \label{eq:label_onehot}\resizebox{0.43\textwidth}{!}{$
y_i \;\;\equiv\;\;(y_i^1,\dots,y_i^K)\,\in\Delta^K_{0,1} \quad 
\text{for}\quad y_i^k := \left[k=y_i\right]$}
\end{equation}
analogously with the hard ground truth labels in \eqref{eq:GT_index} and \eqref{eq:GT_onehot}.
In part, hard pseudo-labels are motivated by the network training where the default 
is NLL loss \eqref{eq:NLL} assuming discrete labels. Besides, powerful discrete solvers 
for the Potts model are well-known \cite{boykov2001interactive,Pock:CVPR2009,chambolle2011first}.
We discuss the potential advantages of soft pseudo-labels in Section \ref{sec:contributions}.

{\bf Overview: joint loss vs ``proposal generation''}:
Most self-labeling approaches can be divided into two groups.  
One group designs pseudo-labeling and the network training sub-problems that
are not formally related, e.g. \cite{lin2016scribblesup,liang2022tree,xu2021scribble}. 
While pseudo-labeling typically depends on the current network predictions and the network fine-tuning
uses such pseudo-labels, the lack of a formal relation between these sub-problems implies that
iterating such steps does not guarantee any form of convergence. Such methods are often referred to as
{\em proposal generation} heuristics.

Alternatively, the pseudo-labeling sub-problem and the network training sub-problem
can be formally derived from a weakly-supervised loss like \eqref{eq:loss_ws}, e.g.
by ADM {\em splitting} \cite{marin2019beyond} or as high-order {\em trust-region} method \cite{marin2021robust}.
Such methods often formulate a {\em joint loss} w.r.t. network predictions and pseudo-labels and iteratively decrease the loss.
These self-labeling methods are mathematically well-founded, easy to understand and reproduce, 
and numerically stable due to guaranteed convergence.

\subsection{Soft pseudo-labels: motivation \& contributions} \label{sec:contributions}

We observe that self-labeling with hard pseudo-labels $y_i$
is inherently limited as such labels cannot represent the uncertainty of class estimates 
at unlabeled pixels $i\in\Omega\backslash S$.  Instead, we focus on {\em soft} pseudo-labels
\begin{equation} \label{eq:label_soft}
y_i \;\;=\;\;(y_i^1,\dots,y_i^K)\,\in\Delta^K
\end{equation}
which are general categorical distributions $p$ over $K$-classes.
It is possible that the estimated pseudo-label $y_i$ in \eqref{eq:label_soft} could be a one-hot distribution, 
which is a vertex of $\Delta^K$. 
In this special case, $y_i$ can be treated as a class index, 
but we use the general formulation \eqref{eq:label_soft} in the main parts of the paper starting Section \ref{sec:our_approach}.
On the other hand, the ground truth labels $\bar{y}_i$ are always hard and we use them either
as indices \eqref{eq:GT_index} or one-hot distributions \eqref{eq:GT_onehot}, as convenient.

Soft pseudo-labels can be found in prior work on weakly-supervised segmentation
\cite{liang2022tree,xu2021scribble} using the ``soft proposal generation''.
In contrast, we formulate soft self-labeling as a principled optimization methodology where
network predictions and soft pseudo-labels are variables in a joint loss, which guarantees convergence 
of the training procedure. Our pseudo-labels are auxiliary variables for ADM-based \cite{boyd2004convex} 
splitting of the loss \eqref{eq:loss_ws} into two simpler optimization sub-problems: 
one focused on the Potts model over unlabeled pixels, and the other on the network training. 
While similar to \cite{marin2019beyond}, instead of hard, we use soft auxiliary variables 
for the Potts sub-problem. Our work can be seen as a study of the relaxed Potts sub-problem 
in the context of weakly-supervised semantic segmentation. The related prior work is
focused on discrete solvers fundamentally unable to represent class estimate uncertainty. 
Our contributions can be summarized as follows:
\begin{itemize}
\item convergent {\em soft self-labeling} framework based on a simple joint self-labeling loss
\item systematic evaluation of Potts relaxations and (cross-) entropy terms in our loss
\item state-of-the-art in scribble-based semantic segmentation based on standard network architectures (no modifications) and well-founded convergent training procedures that are easy to reproduce and generalize to other problems
\item using the same segmentation model, our self-labeling loss with $3\%$ scribbles may outperform standard supervised cross-entropy loss with full ground truth masks.
\end{itemize}


\section{Our soft self-labeling approach} \label{sec:our_approach}


First, we apply ADM splitting \cite{boyd2004convex} to weakly supervised loss \eqref{eq:loss_ws} 
to formulate our self-labeling loss \eqref{eq:loss_self} incorporating additional soft auxiliary variables,
i.e. pseudo-labels \eqref{eq:label_soft}. 
It is convenient to introduce pseudo-labels $y_i$ on all pixels in $\Omega$ even though a subset of pixels (seeds)
$S\subset\Omega$ have ground truth labels $\bar{y}_i$. We will simply impose a constraint 
that pseudo-labels and ground truth labels agree on $S$. Thus, we assume the following set of pseudo-labels
 $$Y_\Omega:=\{y_i\in\Delta^K\,|\,i\in\Omega,\;\text{s.t.}\;y_i =\bar{y}_i\;\text{for}\;i\in S\}.$$
We split the terms in \eqref{eq:loss_ws} into two groups: one includes NLL and entropy $H$ terms keeping 
the original prediction variables $\sigma_i$ and the other includes the Potts relaxation $P$ replacing $\sigma_i$ 
with auxiliary variables $y_i$. This transforms loss \eqref{eq:loss_ws} into expression
$$-\sum_{i\in S} \ln\sigma_i^{\bar{y}_i}  \;+\;\eta\sum_{i\not\in S} H(\sigma_i)
\;+\;\lambda\sum_{ij\in {\cal N}} P(y_i,y_j) $$
equivalent to \eqref{eq:loss_ws} assuming equality $\sigma_i=y_i$. The standard approximation is to
incorporate constraint $\sigma_i\approx y_i$ directly into the loss, e.g. using $KL$-divergence. 
For simplicity, we use weight $\eta$ for $KL(\sigma_i,y_i)$ to combine it with $H(\sigma_i)$ 
into a single cross-entropy term
{\footnotesize \begin{align} \nonumber
-\sum_{i\in S} \ln\sigma_i^{\bar{y}_i} & +\;\underbrace{\eta\sum_{i\not\in S} H(\sigma_i)+\eta\sum_{i\not\in S} KL(\sigma_i,y_i)}\;+\;\lambda\sum_{ij\in {\cal N}} P(y_i,y_j) \\
\nonumber
  & \quad\quad\quad\quad\quad  \eta\sum_{i\not\in S} H(\sigma_i,y_i)
\end{align}}
defining auxiliary {\em self-labeling loss} w.r.t. both $\sigma_i$ and  $y_i$ 
\begin{equation} \label{eq:loss_self}\resizebox{.43\textwidth}{!}{$
-\sum_{i\in S} \ln\sigma_i^{\bar{y}_i}\;+\;\eta \sum_{i\not\in S} H(\sigma_i,y_i)
\;+\;\lambda\sum_{ij\in {\cal N}} P(y_i,y_j)$}
\end{equation}
approximating the original weakly supervised loss \eqref{eq:loss_ws}. 

Iterative minimization of this loss
w.r.t. predictions $\sigma_i$ (model parameters training) and pseudo-labels $y_i$ effectively breaks the original
optimization problem for \eqref{eq:loss_ws} into two simpler sub-problems, assuming there is a good solver for
optimal pseudo-labels. The latter is plausible since the unary term $H(\sigma_i,y_i)$ is convex for $y_i$, and the Potts model (incl. convex and non-convex relaxations) is widely studied.

Section \ref{sec:Potts} discusses standard and new relaxations of the Potts model $P$. 
Section \ref{sec:CE} discusses several robust variants of the cross-entropy term $H$ in \eqref{eq:loss_self}
connecting predictions with uncertain (soft) pseudo-labels $y_i$ at unlabeled points $i\in\Omega \backslash S$. To systematically evaluate these variants of \eqref{eq:loss_self} in Section \ref{sec: experiments}, we propose a general efficient GPU-friendly solver for the corresponding pseudo-labeling sub-problem, which is detailed in the Supplementary Material.

\subsection{Second-order relaxations of the Potts model} \label{sec:Potts}

We limit our focus to basic second-order relaxations and several motivated extensions. We start from two important standard cases (see Table \ref{tab:Potts_relax}): 
{\em quadratic}, the simplest convex relaxation popularized by the {\em random walker} algorithm \cite{grady2006random}, and {\em bi-linear}, which is non-convex but {\em tight} \cite{Ravikumar:ICML2006} w.r.t. the original discrete Potts model. The latter implies that
optimizing it over relaxed variables should lead to a solution consistent with a discrete Potts solver, 
e.g. {\em graph cut} \cite{boykov2001interactive} regularizing the
geometric boundary of the segments \cite{boykov2003geodesic}. The quadratic relaxation produces a significantly different soft solution motivated by random walk probabilities \cite{grady2006random}.
\begin{table}[t]
    \centering
    \resizebox{.4\textwidth}{!}{
    \begin{tabular}{|c|c|} 
       \hline {\bf bi-linear} $\sim$ ``graph cut'' &  {\bf quadratic} $\sim$ ``random walker'' \\ 
       $\quad P_\BL(p,q) \;\; :=\;\; 1 - \;\; p^\top q\quad\quad$  &  $\quad P_\Q(p,q) \;\;:=\;\;\frac{1}{2}\|p-q\|^2 \quad $ \\ & \\ \hline \multicolumn{2}{|c|}{\bf normalized quadratic} \\
       \multicolumn{2}{|c|}{ $\quad P_\NQ(p,q) \;\;:=\;\; 1 - \frac{p^\top q}{\|p\| \|q\|}  
\;\;\quad\quad\quad\equiv\quad\quad\quad\;\;\frac{1}{2}\left\| \frac{p}{\|p\|}- \frac{q}{\|q\|} \right\|^2 \quad $} \\ \multicolumn{2}{|c|}{} \\ \hline
    \end{tabular}}
    \caption{Second-order Potts relaxations, see Fig.\ref{Fig:Prelax}(a,b,c)}
    \label{tab:Potts_relax} \vspace{-3ex}
\end{table}

\begin{figure}[b!]
    \centering
    \resizebox{0.4\textwidth}{!}{
    \begin{tabular}{cc}
         \includegraphics[width=0.45\linewidth]{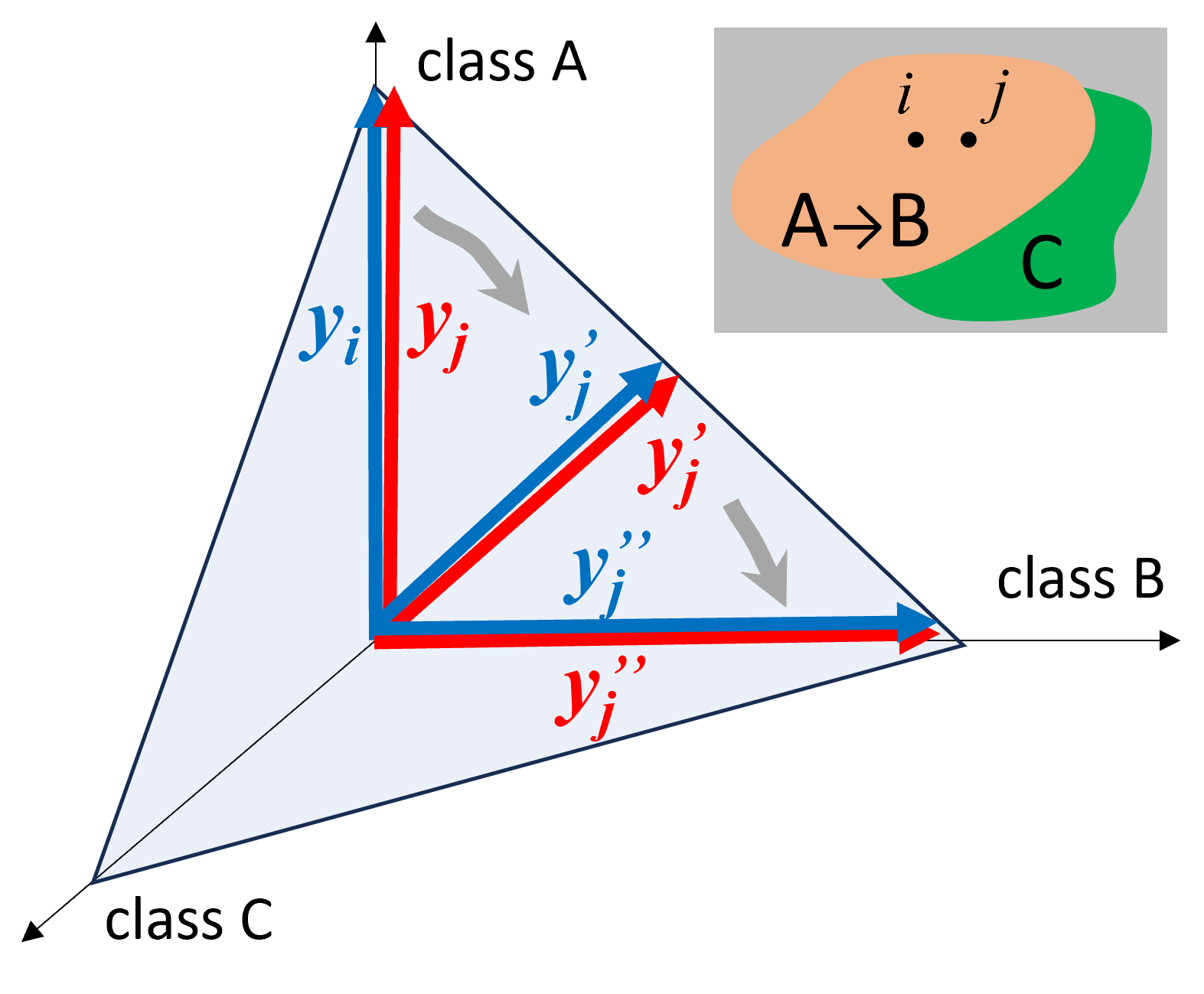} &
          \includegraphics[width=0.45\linewidth]{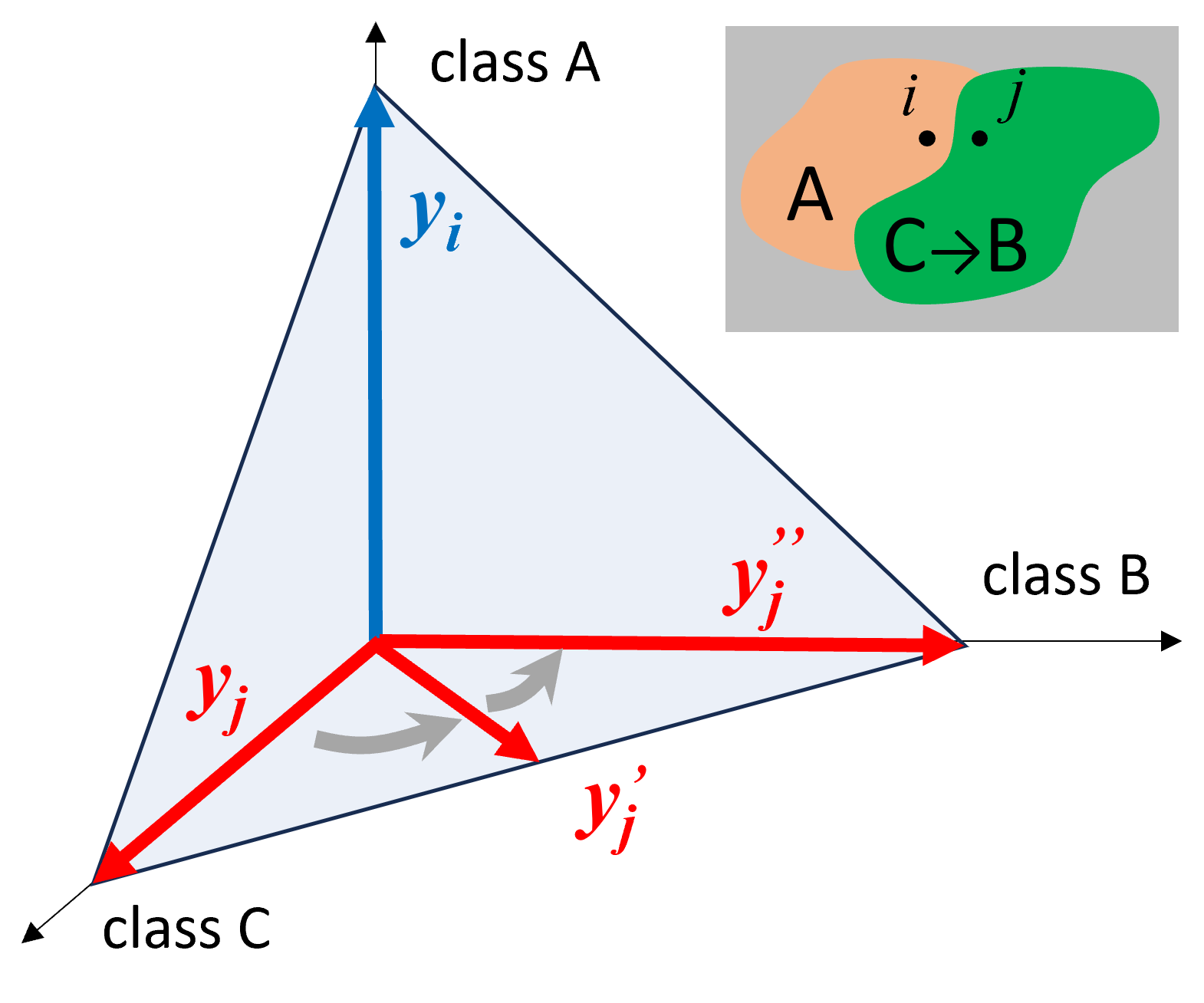} \\
          (a) both pixels change (A $\rightarrow$ B)   & (b) only pixel $j$ changes (C $\rightarrow$ B)
    \end{tabular}}
    \caption{Examples of "moves" for neighboring pixels $\{i,j\}\in{\cal N}$. Their (soft) pseudo-labels $y_i$ and $y_j$ are illustrated on the probability simplex $\Delta^K$ for $K=3$. In (a) both pixels $i$ and $j$ are inside a region/object changing its label from A to B. In (b) pixels $i$ and $j$ are on the boundary between two regions/objects; one is fixed to class A and the other changes from class C to B. }
    \label{Fig:Potts_diagrams}
\end{figure}

While there are specialized solvers for bi-linear and quadratic relaxations, we are interested in a general efficient pseudo-label solver integrated into the standard GPU-based network optimization framework.
This suggests gradient descent as an optimizer for pseudo-labels $y$ (see Suppl. Mat).
Figure \ref{Fig:Potts_diagrams} shows two representative local minima issues
that gradient descent may encounter for (a) the bi-linear and (b) quadratic relaxations of the Potts loss. 
These examples indicate that gradient descent for these relaxations can distort the
properties of the original (discrete) Potts loss.

In (a) two neighboring pixels jointly change the common label from $y_i=y_j=(1,0,0)$ to
$y''_i=y''_j=(0,1,0)$, which corresponds to a ``move'' where the whole object is reclassified from A to B. 
This move does not violate the smoothness constraint represented by the original (discrete) Potts model. But, bilinear relaxation prevents move (a) since the
intermediate state $y'_i=y'_j=(\frac{1}{2},\frac{1}{2},0)$ has a higher cost 
$$P_\BL(y'_i,y'_j) = \frac{1}{2}\;\;> \;\;0 = P_\BL(y_i,y_j) = P_\BL(y''_i,y''_j)$$
while quadratic relaxation assigns constant (zero) loss for all states during this move. 

Figure \ref{Fig:Potts_diagrams}(b) shows a move problematic for the quadratic
relaxation. Two neighboring pixels on the boundary of objects A and C have labels $y_i=(1,0,0)$ and $y_j=(0,0,1)$. 
The second object
attempts to change from class label C to B that does not affect the Potts discontinuity penalty between two pixels.
But, quadratic relaxation prefers that the second object stays in the intermediate soft state $y'_j=(0,\frac{1}{2},\frac{1}{2})$
$$P_\Q(y_i,y'_j) = \frac{3}{4}\;\;< \;\;1 = P_\Q(y_i,y_j) = P_\Q(y_i,y''_j)$$
while bi-linear relaxation $P_\BL(y_i,y_j)=1$ remains constant as $y_j$ transitions from state C to B.

\begin{figure*}[t]
    \centering
    \resizebox{\textwidth}{!}{\begin{tabular}{c|cc|cc|cc}
         \multirow{2}{*}[2em]{\parbox{3mm}{\rotatebox[origin=c]{90}{Table \ref{tab:Potts_relax}}} } &
         \includegraphics[width=0.2\linewidth]{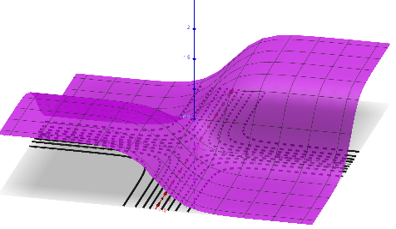} &
         \includegraphics[width=0.1\linewidth]{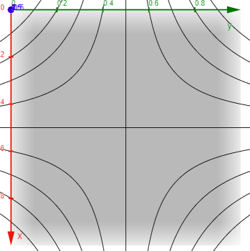} &
         \includegraphics[width=0.2\linewidth]{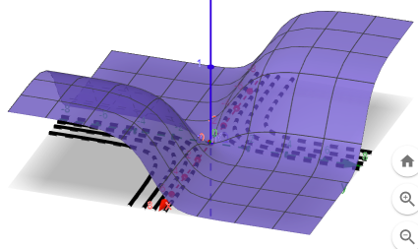} &
         \includegraphics[width=0.1\linewidth]{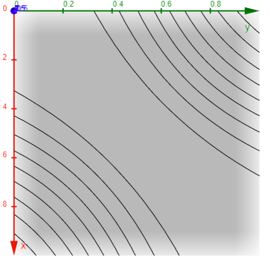} &
          \includegraphics[width=0.2\linewidth]{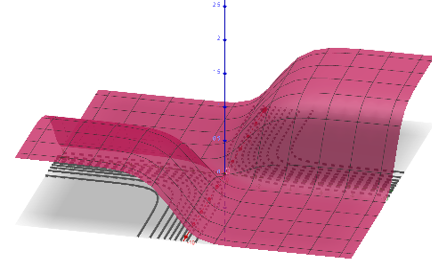} &
          \includegraphics[width=0.1\linewidth]{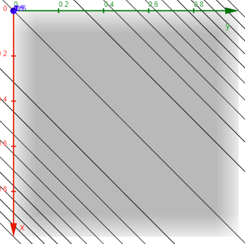} \\
          & \multicolumn{2}{c|}{(a) bi-linear $P_\BL$} & \multicolumn{2}{c|}{(b) normalized quadratic $P_\NQ$} & 
          \multicolumn{2}{c}{(c) quadratic $P_\Q$}  \\ \hline
          \multirow{2}{*}[5em]{\parbox{3mm}{\rotatebox[origin=c]{90}{Table \ref{tab:Potts_relax_log} \hspace{0.5ex} (+log)}}} &
         \includegraphics[width=0.2\linewidth]{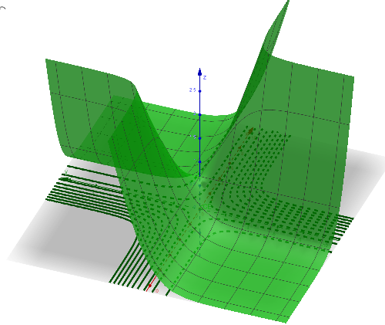} &
         \includegraphics[width=0.1\linewidth]{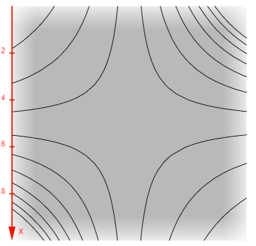} &
         \includegraphics[width=0.2\linewidth]{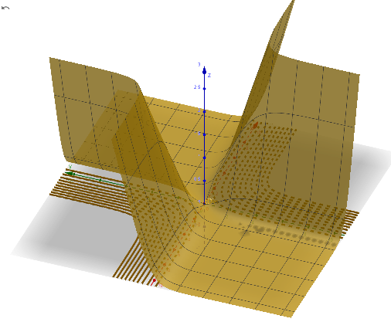} &
         \includegraphics[width=0.1\linewidth]{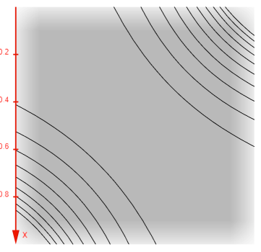} &
          \includegraphics[width=0.2\linewidth]{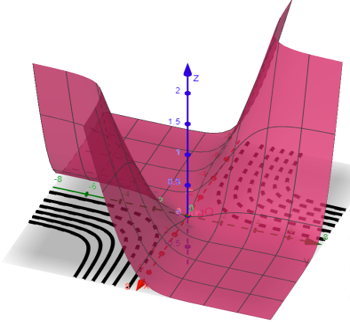} &
          \includegraphics[width=0.1\linewidth]{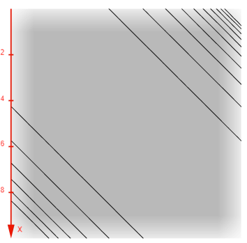} \\
          & \multicolumn{2}{c|}{(d) collision cr. entropy $P_\CCE$} & \multicolumn{2}{c|}{(e) collision divergence $P_\CD$} & 
          \multicolumn{2}{c}{(f) log-quadratic $P_\LQ$}          
    \end{tabular}}
    \caption{Second-order Potts relaxations in Tables \ref{tab:Potts_relax} and \ref{tab:Potts_relax_log}: interaction potentials $P$ for pairs of predictions $(\sigma_i,\sigma_j)$ in \eqref{eq:loss_ws} or pseudo-labels $(y_i,y_j)$ in \eqref{eq:loss_self} are illustrated for $K=2$ when each $\sigma_i$ or $y_i$ 
    (binary distributions in $\Delta^2$) can be represented by a single scalar as $(x,1-x)$. 
    The contour maps are iso-levels of $P((x_i,1-x_i),(x_j,1-x_j))$ over domain $(x_i,x_j)\in [0,1]^2$.
    The 3D plots above illustrate the potentials $P$ as functions over pairs of ``logits'' $(l_i,l_j)\in{\mathbb R}^2$ where each scalar $l_i$ defines binary distribution $(x_i,1-x_i)$ for $x_i = \frac{1}{1+e^{-2l_i}}\in[0,1]$.}
    \label{Fig:Prelax}
\end{figure*}

We propose a new relaxation, {\em normalized quadratic} in Table \ref{tab:Potts_relax}.
Normalization leads to equivalence between quadratic and bi-linear formulations combining 
their benefits. As easy to check, normalized quadratic relaxation $P_\NQ$ 
removes the local minima in both examples of Figure \ref{Fig:Potts_diagrams}. 
Table \ref{tab:Potts_relax_log} also proposes ``logarithmic'' versions of the relaxations in  Table \ref{tab:Potts_relax} composing them with function $-\ln(1-x)$. 
As illustrated by Figure \ref{Fig:Prelax}, the logarithmic versions (d-f) address the ``vanishing gradients'' evident from the flat regions in (a-c). 

\begin{table}[t]
    \centering
    \resizebox{0.4\textwidth}{!}{
    \begin{tabular}{|c|c|} 
       \hline {\bf collision cross entropy} &  {\bf log-quadratic} \\ 
       $\quad P_\CCE(p,q) \;\; :=\;\; -\ln p^\top q  \quad\quad$  &  $\quad P_\LQ(p,q) \;\;:=\;\;-\ln \left(1-\frac{\|p-q\|^2}{2}\right)  \quad $ \\ & \\ \hline \multicolumn{2}{|c|}{\bf collision divergence} \\
       \multicolumn{2}{|c|}{ $\quad P_\CD(p,q) \;\;:=\;\; -\ln \frac{p^\top q}{\|p\| \|q\|}    
\;\;\quad\quad\quad\equiv\quad\quad\quad -\ln\left(1-\frac{1}{2}\left\| \frac{p}{\|p\|} - 
\frac{q}{\|q\|}  \right\|^2 \right) \quad $} \\ \multicolumn{2}{|c|}{} \\ \hline
    \end{tabular}}
    \caption{Log-based Potts relaxations, see Fig.\ref{Fig:Prelax}(d,e,f)}
    \label{tab:Potts_relax_log} \vspace{-3ex}
\end{table}


\subsection{Cross-entropy and soft pseudo-labels} \label{sec:CE}
\begin{figure*}
    \centering
    \begin{tabular}{c@{\hspace{1\tabcolsep}}c@{\hspace{1\tabcolsep}}c@{\hspace{1\tabcolsep}}c}
         \includegraphics[width=0.2\linewidth,]{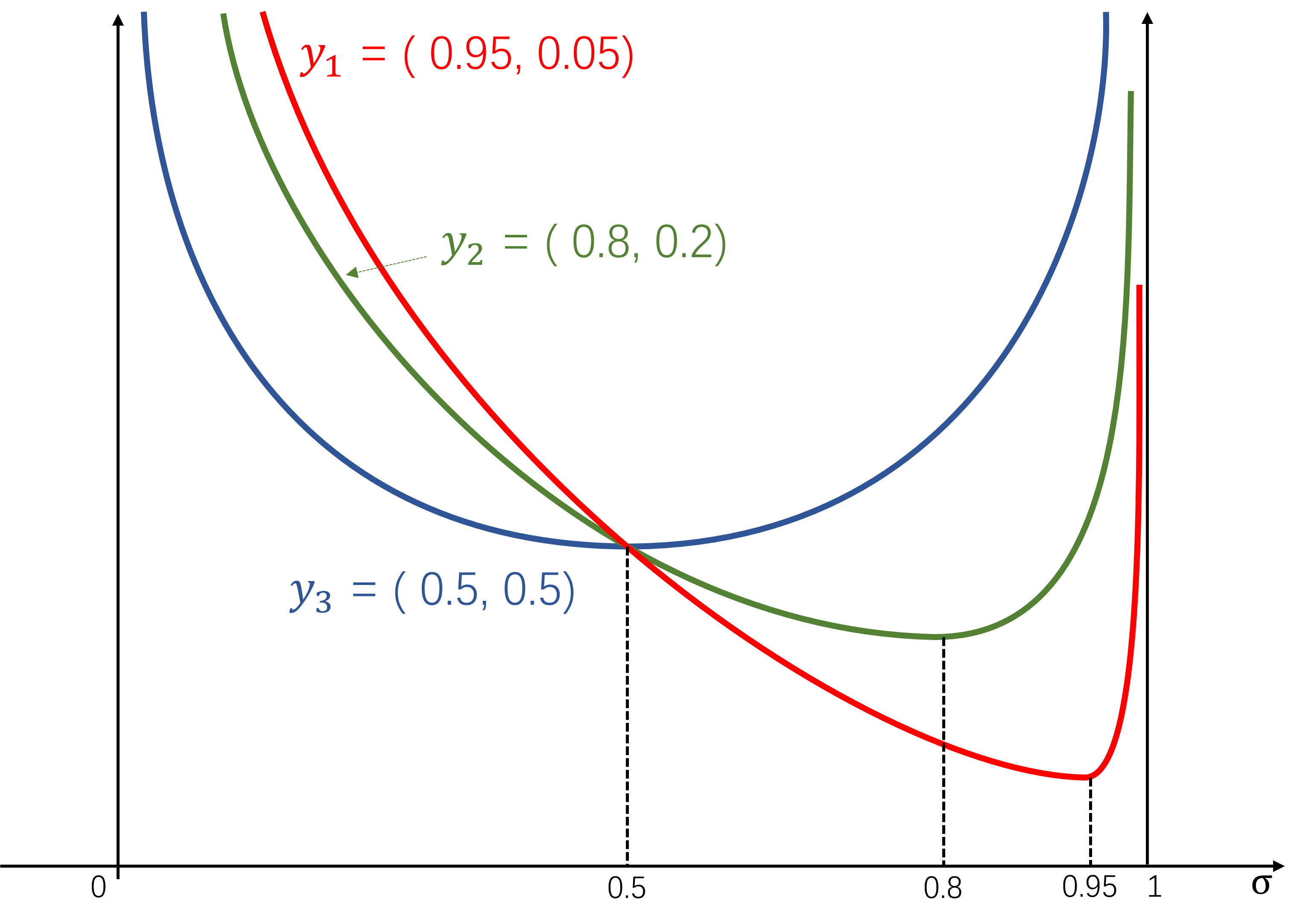} &
         \includegraphics[width=0.2\linewidth,]{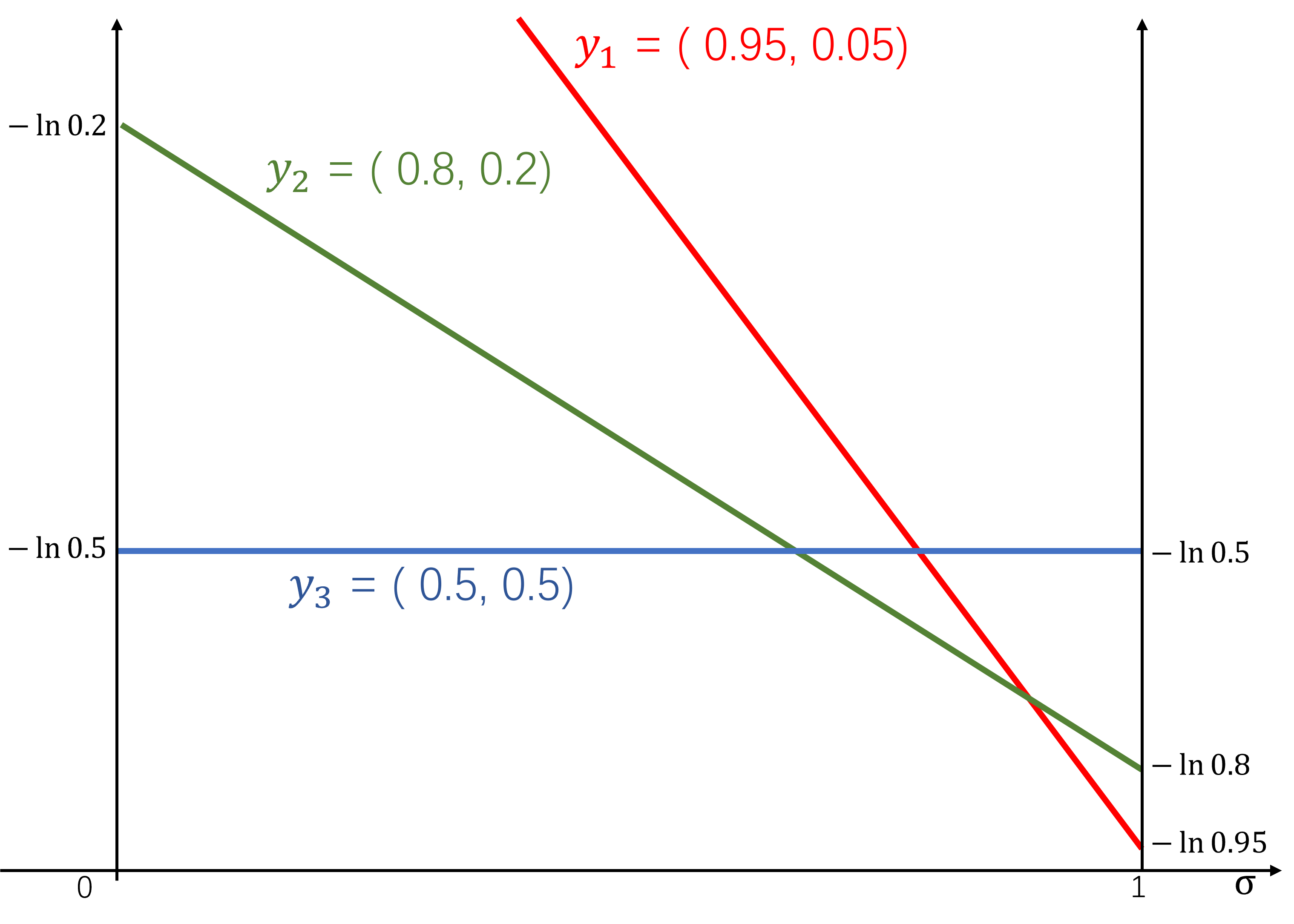} &
         \includegraphics[width=0.2\linewidth,]{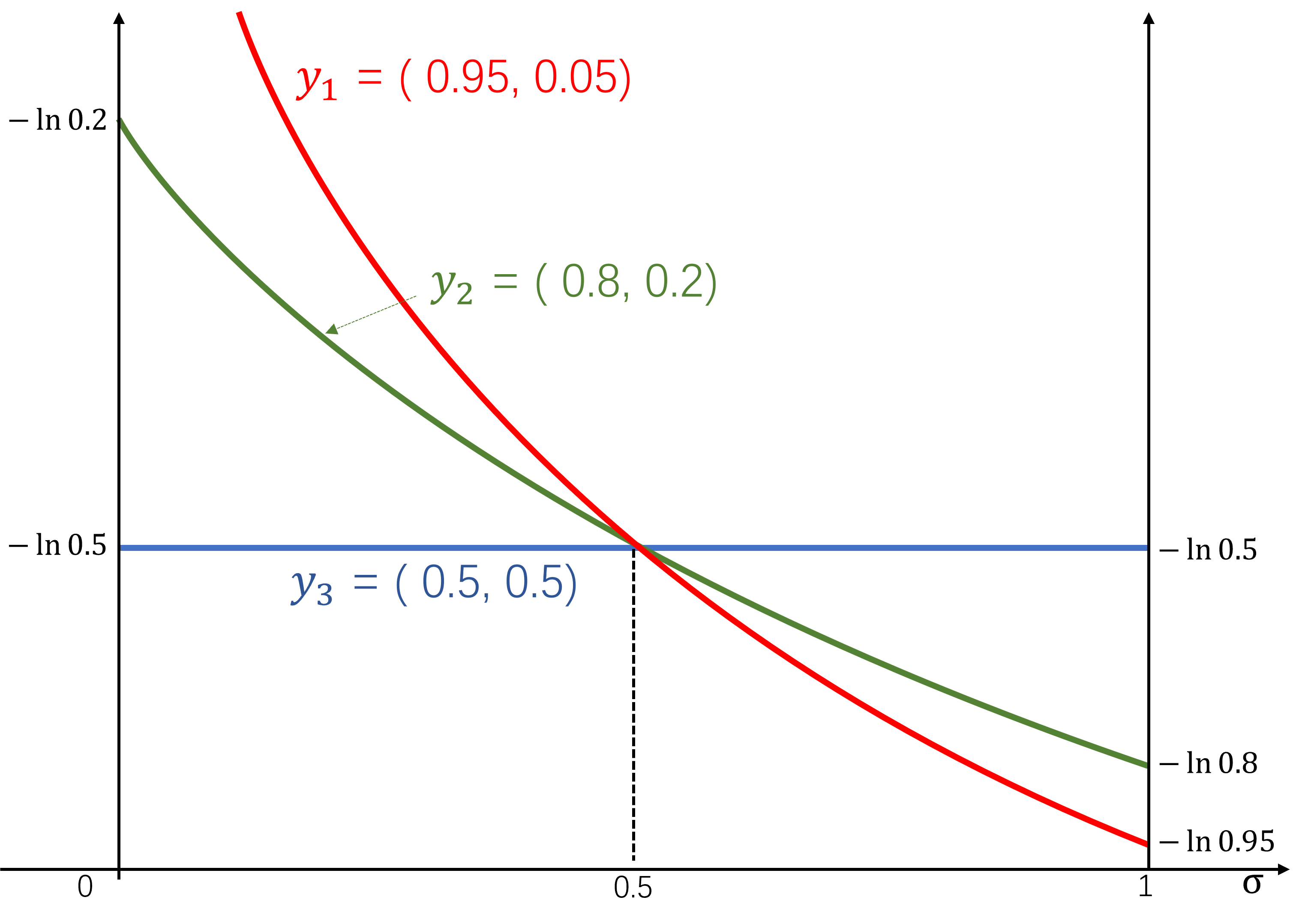} &
         \includegraphics[width=0.3\linewidth, height=2.5cm]{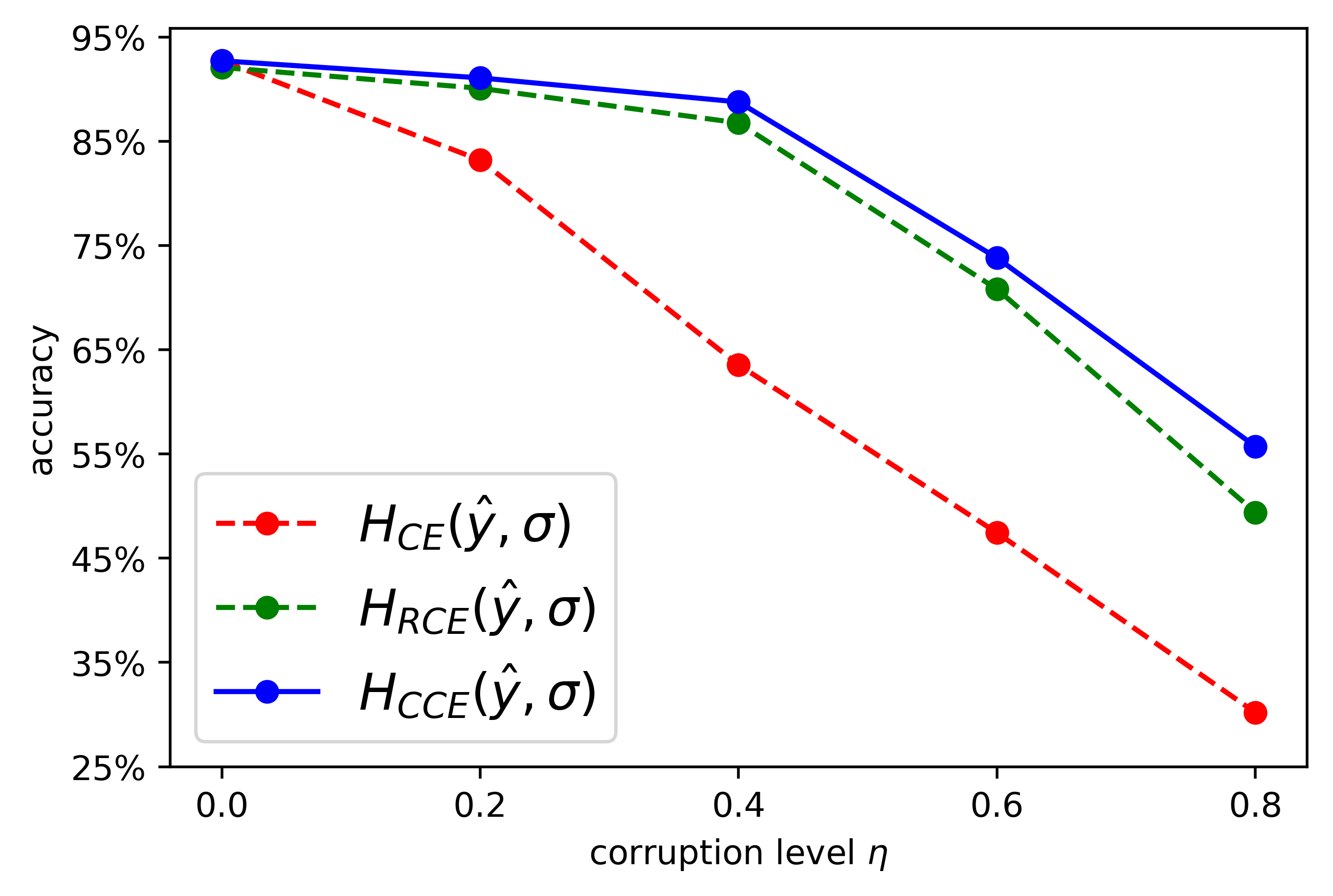}\\
          (a) standard $H_\CE(y, \sigma)$ & (b) reverse $H_\RCE (y, \sigma)$ & (c) collision $H_\CCE(y, \sigma)$ & (d) {\small empirical comparison}
    \end{tabular}
    \caption{Illustration of cross-entropy functions: (a) standard \eqref{eq:D_CE}, (b) reverse \eqref{eq:D_reverseCE}, and (c) collision \eqref{eq:D_cCE}. (d) shows the empirical comparison on the robustness to label uncertainty. The test uses ResNet-18 architecture on fully-supervised {\em Natural Scene} dataset \cite{NSD} 
    where we corrupted some labels. The horizontal axis shows the percentage $\eta$ of training images where 
    the correct ground truth labels were replaced by a random label. All losses trained 
    the model using soft target distributions $\hat{y}=\eta*u+(1-\eta)*y$ representing the mixture of
    one-hot distribution $y$ for the observed corrupt label and 
    the uniform distribution $u$, following \cite{muller2019does}. 
    The vertical axis shows the test accuracy. Training with the reverse and collision cross-entropy is robust 
    to much higher levels of label uncertainty.}
    \label{Fig: cross-entropy term}
\end{figure*}
Shannon's cross-entropy $H(y,\sigma)$ is the most common loss for training network predictions 
$\sigma$ from ground truth labels $y$ in the context of classification, semantic segmentation, etc. 
However, this loss may not be ideal for applications where the targets $y$ are soft categorical distributions 
representing various forms of class uncertainty. For example, this paper is focused on scribble-based segmentation where 
the ground truth is not known for most of the pixels, and the network training is
done jointly with estimating {\em pseudo-labels} $y$ for the unlabeled pixels. 
In this case, soft labels $y$ are distributions representing class uncertainty. 
We observe that if such $y$ is used as a target in $H(y,\sigma)$, the network 
is trained to reproduce the uncertainty, see Figure~\ref{Fig: cross-entropy term}(a). This motivates
the discussion of alternative ``cross-entropy'' functions where the quotes indicate an 
informal interpretation of this information-theoretic concept. Intuitively, such functions should
encourage decisiveness, as well as proximity between the predictions and pseudo-labels, but avoid mimicking the uncertainty in both directions:
from soft pseudo-labels to predictions and vice-versa.
We show that the last property can be achieved in a probabilistically principled manner.
The following paragraphs discuss three cross-entropy functions $H$ that we study in the context of our
self-labeling loss \eqref{eq:loss_self}.

{\bf Standard cross-entropy} provides the obvious baseline for evaluating two alternative versions that follow.
For completeness, we include its mathematical definition
\begin{equation} \label{eq:D_CE}
H_\CE(y_i,\sigma_i)\;\;=\;\;H(y_i,\sigma_i) \;\;\equiv\;\;-\sum_k y_i^k \ln \sigma_i^k 
\end{equation}
and remind the reader that this loss is primarily used with hard or one-hot labels, in which case it is also equivalent to NLL loss $-\ln\sigma^{y_i}_i$
previously discussed for ground truth labels \eqref{eq:NLL}. As mentioned earlier, Figure \ref{Fig: cross-entropy term}(a) shows that for soft pseudo-labels like
$y=(0.5,0.5)$, it forces predictions to mimic or replicate the uncertainty $\sigma\approx y$. In fact, label $y=(0.5,0.5)$ just tells that the class is unknown and
the network should not be supervised by this point. This problem manifests itself in the poor performance of the standard cross-entropy \eqref{eq:D_CE}  in our experiment discussed in
Figure~\ref{Fig: cross-entropy term} (d) (red).

{\bf Reverse cross-entropy} is defined in this paper as 
\begin{equation} \label{eq:D_reverseCE}
H_\RCE(y_i,\sigma_i)\;\;=\;\;H(\sigma_i,y_i)\;\;\equiv\;\;-\sum_k \sigma_i^k \ln y_i^k
\end{equation}
switching the order of labels and predictions in \eqref{eq:D_CE}, which is not common. Indeed, Shannon's cross-entropy is not symmetric and the first argument is normally the {\em target} distribution and the second is 
the {\em estimated} distribution. However, in our case, both distributions are estimated and there is no reason not to try the reverse order.
It is worth noting that our self-labeling formulation \eqref{eq:loss_self} suggests that reverse cross-entropy naturally appears when the ADM approach splits the decisiveness and fairness into separate sub-problems.
Moreover, as Figure \ref{Fig: cross-entropy term}(b) shows, in this case, the network does not mimic uncertain pseudo-labels, e.g. the gradient of the blue line is zero.
The results for the reverse cross-entropy in Figure~\ref{Fig: cross-entropy term} (d) (green) are significantly better than for the standard (red).
Unfortunately, now pseudo-labels $y$ mimic the uncertainty in predictions $\sigma$.

    

{\bf Collision cross-entropy} defined in this paper as
\begin{equation} \label{eq:D_cCE}
H_\CCE(y_i,\sigma_i)\;\;\equiv\;\;-\ln \sum_k \sigma_i^k y_i^k \;\;\equiv\;\;-\ln \sigma^\top y    
\end{equation}
resolves the problem in a principled way. It is symmetric w.r.t. pseudo-labels and predictions. 
The dot product $\sigma^\top y$ can be seen as a probability of equality of two random variables, 
the predicted class $C$ and unknown true class $T$, which are represented by the distributions $\sigma$ and $y$.
Indeed, $$\Pr(C=T) = \sum_k Pr(C=k)\Pr(T=k) = \sigma^\top y.$$ 
Loss \eqref{eq:D_cCE} maximizes this ``collision'' probability instead of enforcing the equality of 
distributions $\sigma= y$.
Figure \ref{Fig: cross-entropy term}(c) shows no mimicking of uncertainty (blue line). Unlike the reverse cross-entropy, this is also
valid when $y$ is estimated from uncertain predictions $\sigma$ since \eqref{eq:D_cCE} is symmetric. This leads to the best performance in
Figure~\ref{Fig: cross-entropy term} (d) (blue). Our extensive experiments are conclusive that collision cross-entropy 
is the best option for $H$ in self-labeling loss \eqref{eq:loss_self}.

\section{Experiments}
\label{sec: experiments}
\begin{figure}[h]
    \centering
    \resizebox{0.4\textwidth}{!}{\begin{tabular}{ccc}
        \includegraphics[width=0.4\linewidth]{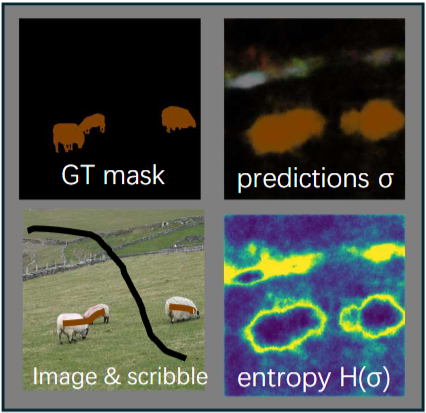}  &
        \multicolumn{2}{c}{\includegraphics[width=0.6\linewidth]{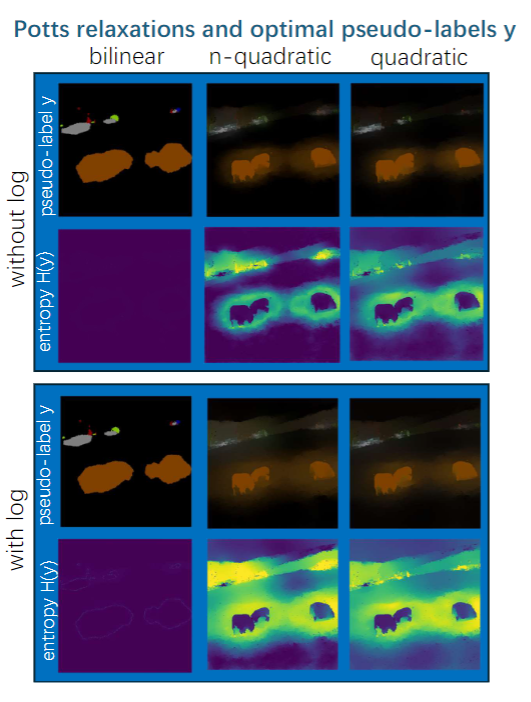}}\\
        (a) Image, GT \& input & \\[-2.8ex]
         & \multicolumn{2}{c}{(b) Pseudo-labels using different Potts relaxation}\\
    \end{tabular}}
    \caption{Illustration of the difference among Potts relaxations. The visualization of soft pseudo-labels uses the convex combination of RGB colors for each class weighted by pseudo-label itself.}
    \label{Fig: pairwise illustration}
\end{figure}

Our experiments evaluate the components of our loss \eqref{eq:loss_self} (cross-entropy, pairwise term, and neighborhood). We compare standard segmentation architectures trained using the ``textbook'' approach, 
i.e. loss minimization, to the state-of-the-art methods, which often require network modifications 
and ad-hoc training procedures. 
Our quantitative tests in Section \ref{sec: comparison potts relaxation} evaluate different Potts relaxations. 
Some qualitative examples are presented in Figure~\ref{Fig: pairwise illustration}. Section \ref{sec: comparison cross entropy terms} compares three cross-entropy terms. Section \ref{sec: experiment neighborhood system} 
evaluates our soft self-labeling loss on the nearest-neighbor and dense neighborhood systems. We summarized the results in Section \ref{sec: soft vs hard vs GD}. Section \ref{sec: experiment SOTA} shows 
that our method achieves the SOTA and, in some cases, may outperform the full supervision.

\paragraph{Dataset and evaluation}
We mainly use the standard PASCAL VOC 2012 dataset \cite{everingham2009pascal} and scribble annotations for supervision \cite{lin2016scribblesup}. The dataset contains 21 classes including background. Following the common practice \cite{chen2014semantic,tang2018normalized,tang2018regularized}, we use the augmented version with 10,582 training images and 1449 images for validation. Our evaluation metric is the standard mean Intersection-over-Union (mIoU) on the validation set.
We also test our method on two additional datasets in Section \ref{sec: experiment SOTA}. One is Cityscapes \cite{Cordts2016Cityscapes} which is built for urban scenes and consists of 2975 and 500 fine-labeled images for training and validation. There are 19 out of 30 annotated classes for semantic segmentation.
 The other one is ADE20k \cite{zhou2017scene} which has 150 fine-grained classes. There are 20210 and 2000, images for training and validation. Instead of scribble-based supervision, we followed \cite{liang2022tree} to use the block-wise annotation as a form of weak supervision.

\paragraph{Implementation details}
We use DeepLabv3+ \cite{chen2018encoder} with two backbones: ResNet101 \cite{he2016deep} and MobileNetV2 \cite{sandler2018mobilenetv2}. We use ResNet101 in Section \ref{sec: experiment SOTA}, and MobileNetV2 in other sections, for efficiency. We also use Vision Transformer backbone \cite{dosovitskiy2020image}, specifically $\text{vit-base-patch16-224}$, and linear decoder in Section \ref{sec: experiment SOTA}. All backbone networks (ResNet-101 and MobileNetV2, ViT) are pre-trained on Imagenet \cite{deng2009imagenet}. Unless stated explicitly, we use batch 12 as the default across all the experiments. To avoid feeding random predictions from the uninitialized classification head into pseudo-labeling sub-problem, we initialize the network using the cross-entropy on scribbles, i.e. the first term in loss \eqref{eq:loss_self}, similarly to pre-training in \cite{tang2018normalized}. The network parameters are optimized by SGD. The initial learning rate 0.0007 
is scheduled by a polynomial decay of power 0.9. 
All loss variants are optimized for 60 epochs, but all hyperparameters are tuned separately for each variant. 
For our best result, we use $\eta=0.3, \lambda=6$, $H_\CCE$ and $P_\CD$. The color-bandwidth in the Potts model is set to 9 across all the experiments on Pascal VOC 2012 and 3 for Cityscapes and ADE20k datasets.


\subsection{Comparison of Potts relaxations}
\label{sec: comparison potts relaxation}
To compare different Potts relaxations within our self-labeling framework, we select $H_\CCE$ as the
cross-entropy term in \eqref{eq:loss_self}, as motivated by the results
\begin{table}\centering
\resizebox{0.4\textwidth}{!}{\begin{tabular}{c|ccccc}
\hline
& \multicolumn{5}{c}{scribble length ratio} \\ & 0 & 0.3 & 0.5 & 0.8 & 1.0  \\
\hline
 $P_\BL$ & 56.42& 61.74& 63.81& 65.73 & 67.24 \\
 $P_\NQ$ & 59.01 &65.53&67.80&70.63&71.12 \\
 $P_\Q$ & 58.92& 65.34&67.81&70.43&71.05  \\
 $P_\CCE$ & 56.40& 61.82& 63.81&65.81 & 67.41 \\
  \rowcolor{gray!15}$P_\CD$ & 59.04 & 65.52& 67.84& 70.93& 71.22 \\
 $P_\LQ$ & 59.03& 65.44& 67.81 & 70.80 & 71.21\\
\hline
\end{tabular}}
\caption{Comparison of Potts relaxations with self-labeling. mIoUs on the validation set are shown here.}
\label{tab: comparison pairwise terms}
\end{table}
in Section \ref{sec: comparison cross entropy terms}. The neighborhood system is the nearest neighbors. The quantitative results
are in Table \ref{tab: comparison pairwise terms}. First, one can see that the log-based pairwise terms 
work consistently better. The likely explanation is that the logarithm addresses 
the gradient vanishing problem that is obvious in the top row of Figure \ref{Fig:Prelax}. Moreover, the logarithm encourages smoother transitions across the boundaries (see Figure~\ref{Fig: pairwise illustration}). This may benefit network training when the exact ground truth is unknown. Intuitively, higher uncertainty around object boundaries is natural. Second, the normalized relaxations $P_\NQ$ and $P_\CD$  work better, confirming 
the limitations of $P_\BL$ and $P_\Q$ discussed in Fig.~\ref{Fig:Potts_diagrams}. 

While we focus on a general soft pseudo-label solver (see Suppl.Mat), our results can motivate specialized solvers. 
For example, quadratic relaxation $P_\Q$ seems attractive as it defines a convex loss \eqref{eq:loss_self}. 
If using ``quadratic cross-entropy'' $H_\Q:=\|y-\sigma\|^2$, then our loss even has a closed-form solution for $y$ \cite{grady2005multilabel,grady2006random}. However, the corresponding pseudo-labels tend to be overly soft and additional 
{\em decisiveness} (non-convex entropy) term improves mIoU. 

\subsection{Comparison of cross-entropy terms}
\label{sec: comparison cross entropy terms}
\begin{figure}\centering
    \centering
    \includegraphics[width=0.85\linewidth]{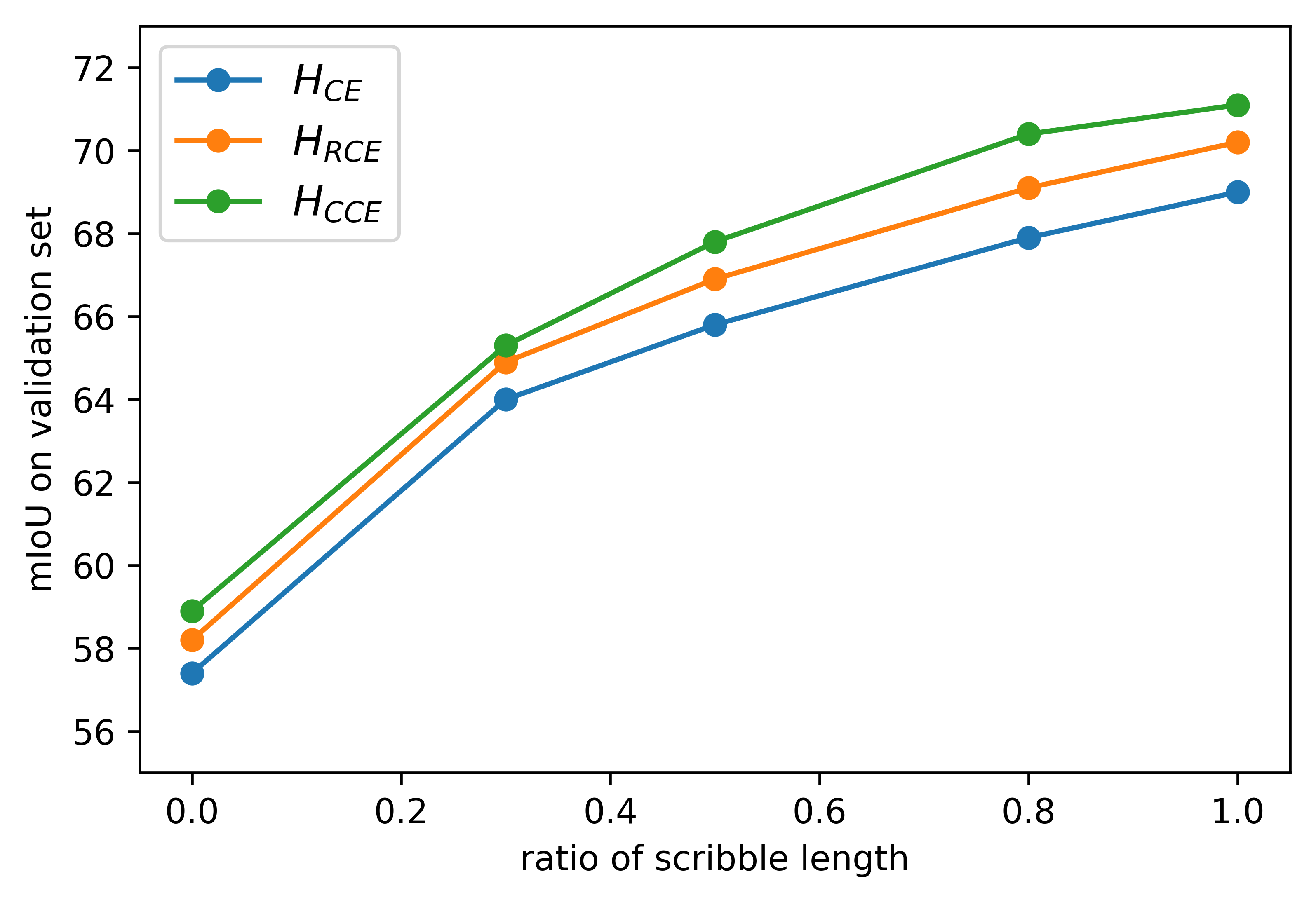}
    \vspace{-2.2ex}
    \caption{Comparison of cross-entropy terms.}
    \label{fig: comparison on data term}
\end{figure}

We compare different cross-entropy terms while fixing the pairwise term to $P_\Q$ due to its simplicity.
We also use the nearest neighbors. The results are in Figure~\ref{fig: comparison on data term}.
In agreement with Fig.\ref{Fig: cross-entropy term}(d), $H_\CCE$ performs consistently better across different
supervision levels, i.e. scribble lengths. Both $H_\CCE$ and $H_\RCE$ have a strong margin over the standard
entropy $H_\CE$ as network training becomes robust to uncertainty and errors in pseudo labels, see Sec.\ref{sec:CE}.


\subsection{Comparison of neighborhood systems}
\label{sec: experiment neighborhood system}
So far we used the four nearest neighbors (NN) neighborhood for the pairwise Potts term in \eqref{eq:loss_self}. 
Here we compare NN with the dense neighborhoods (DN). We use $H_\CCE$ as the cross-entropy term. 
To optimize pseudo-labels over DN, we still use the general gradient descent technique detailed in the Supplementary Material. The gradient computation for DN Potts can employ the efficient {\em bilateral filtering} 
\cite{tang2018normalized}, but it applies only to (proper) second-order variants, i.e. $P_\BL$ and $P_\Q$. 
We focus on $P_\Q$ as a preferred option for soft self-labeling. We obtained 71.1\% mIoU on nearest neighbors while only getting 67.9\% on dense neighborhoods
\begin{figure}
    \centering
   \includegraphics[width=\linewidth]{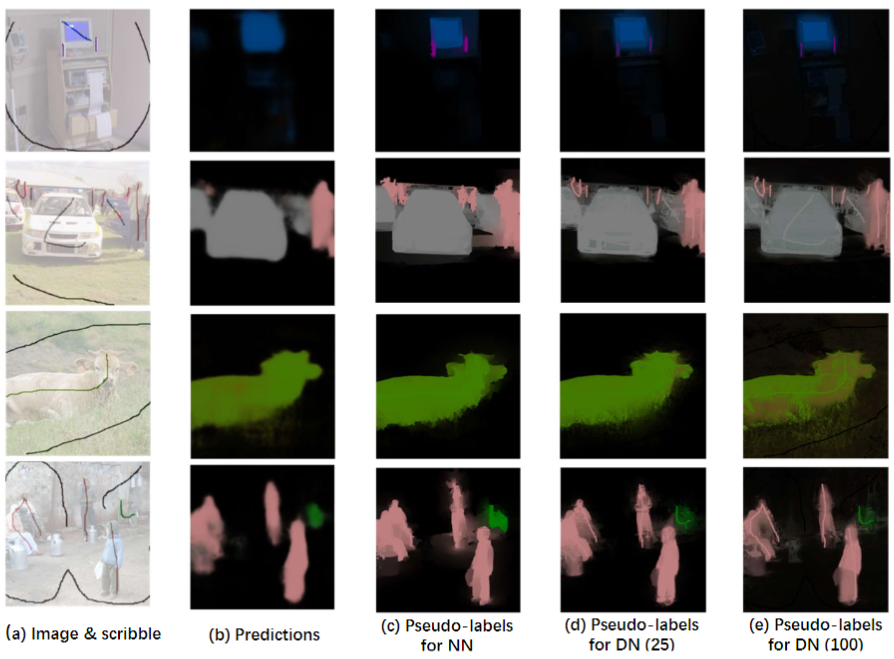}
    \caption{Pseudo-labels generated from given network predictions using different neighborhoods: nearest (NN) and dense (DN).}
    \label{fig: neighborhood comparison}
\end{figure}
(bandwidth 100). Some qualitative results are shown in Figure~\ref{fig: neighborhood comparison} indicating that larger neighborhoods induce lower-quality pseudo-labels. A possible explanation is that the Potts model reduces to the cardinality/volume potentials for larger neighborhoods \cite{veksler2019efficient}. The nearest neighborhood is better for edge alignment producing better weakly-supervised results.



\begin{table}\centering
\resizebox{0.4\textwidth}{!}{\begin{tabular}{|P{1cm}|P{1cm}|P{2.5cm}|P{2.5cm}|}\cline{3-4}
         \multicolumn{2}{c|}{} & \multicolumn{2}{c|}{$\mathcal{N}$} \\\cline{3-4}
        \multicolumn{2}{c|}{} & NN & DN  \\\hline
         \multicolumn{2}{|c|}{GD} & 67.0 & 69.5$^*$ \cite{tang2018regularized}  \\\hline
         \multirow{2}{*}{SL} & hard & 69.6$^*$ \cite{marin2021robust} & 63.1 \cite{lin2016scribblesup} \\\cline{2-4}
         & soft & \textbf{71.1} & 67.9 \\\hline
    \end{tabular}}
\caption{Summary of Potts-based WSSS (full-scribbles, DeepLab V3+). Reproduced results using public code are marked by $*$.}\label{tab: soft sl vs hard sl vs gd}
\end{table} 
\begin{figure}\centering
    \includegraphics[width=0.75\linewidth]{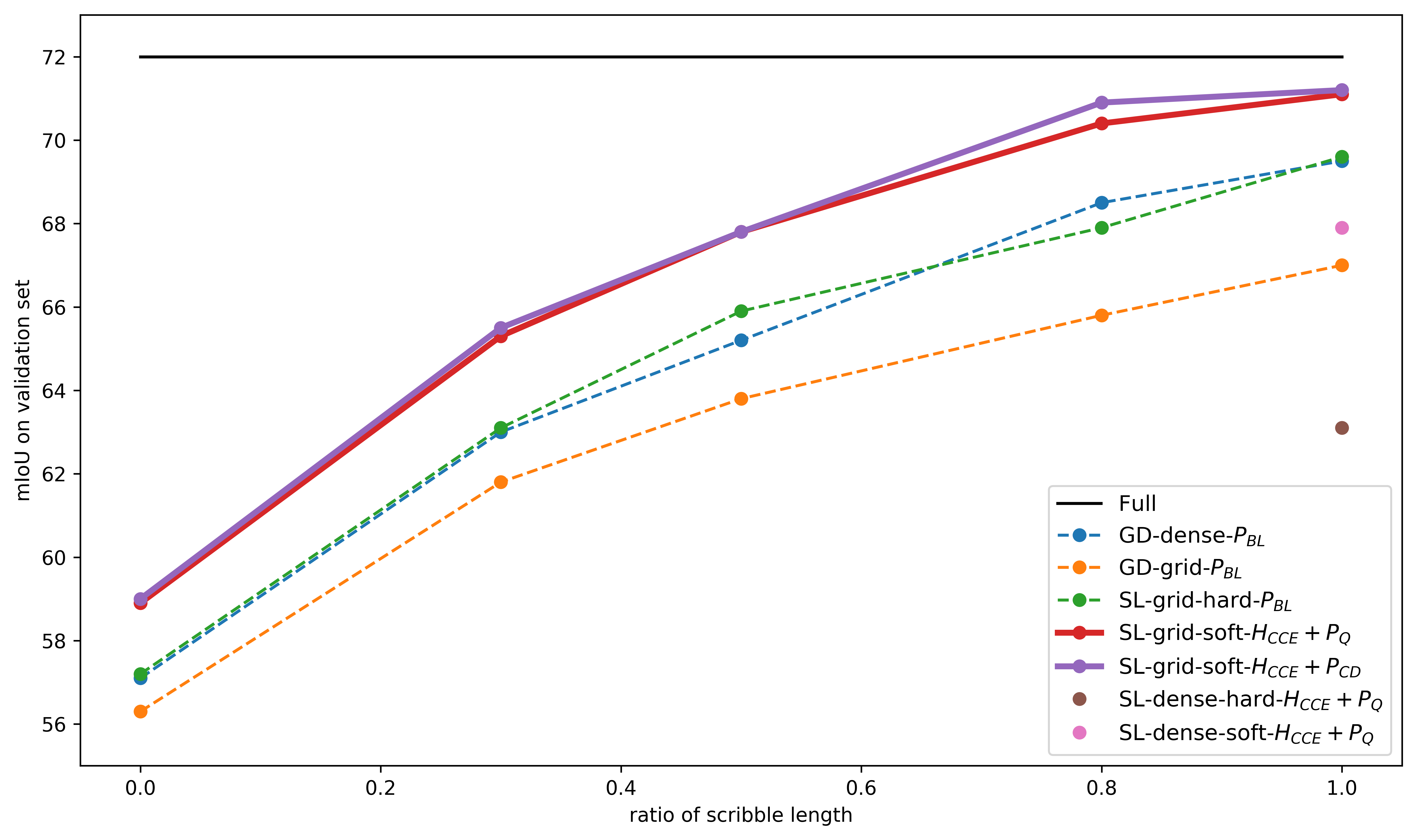}
    \caption{Comparison of different methods using Potts relaxations. The architecture is DeeplabV3+ with the backbone MobileNetV2.}
    \label{fig: soft sl vs hard sl vs gd}
\end{figure}
\begin{table*}[th]
    \centering
    \resizebox{.85\textwidth}{!}{
    \begin{tabular}{P{4.5cm}|P{2.5cm}|P{1.5cm}|P{1cm}|P{1cm}|P{1cm}|P{2cm}|P{1.5cm}}
    \hline
    \multirow{3}{*}{Method} & \multirow{3}{*}{Architecture} &
    \multirow{3}{*}{Batchsize} & \multicolumn{3}{c|}{Training Algorithm} & \multirow{3}{*}{$\mathcal{N}$}  &
    \multirow{3}{*}{mIoU} \\\cline{4-6}
      &  & & GD &\multicolumn{2}{c|}{SL} &  &  \\\cline{5-6}
     &  & &  & hard & soft &  & \\\hline\hline
     \multicolumn{8}{c}{\textbf{Full supervision}: standard architectures + full masks training (NLL loss) }\\\hline
     Deeplab \cite{chen2017deeplab} & V2 & 12 & \checkmark & - & - & - & 75.6\\
     Deeplab$^*$ \cite{chen2018encoder} & V3+ & 12 & \checkmark & - & - & - & \textcolor{teal}{76.6} \\
     Deeplab$^*$ \cite{chen2018encoder} & V3+ & 16 & \checkmark & - & - & - & 78.9 \\
     Vision Transformer \cite{dosovitskiy2020image} & ViT-linear & 16 & \checkmark & - & - & - & 81.4 \\\hline\hline
     \multicolumn{8}{c}{\textbf{Scribble supervision} (full-length scribbles)}\\\hline
     \multicolumn{8}{c}{Modified architectures ($\triangle$), and/or ad-hoc self-labeling ($\circ$) } \\ \hline
     BPG \cite{wang2019boundary} & V2 ($\triangle$) & 10 & \checkmark & - & - & - & 73.2 \\
     URSS \cite{pan2021scribble} & V2 ($\triangle$) & 16 & \checkmark & - & - & - & 74.6 \\ 
     SPML \cite{ke2021universal} & V2 ($\triangle$) & 16 & \checkmark & - & - & - & 74.2 \\
     PSI \cite{xu2021scribble} & V3+ ($\triangle$) & - & - & - & \checkmark ($\circ$) & - & 74.9 \\
     SEMINAR \cite{chen2021seminar} & V3+ ($\triangle$) & 12 & \checkmark & - & - & - & 76.2\\
     TEL \cite{liang2022tree} & V3+ & 16 & - & - & \checkmark ($\circ$) & - & 77.1 \\
     AGMM \cite{wu2024modeling} & ViT-linear ($\triangle$) & 16 & - & - & \checkmark ($\circ$) & - & 78.7 \\
     \hline
     \multicolumn{8}{c}{Standard architectures + Potts loss optimization (relaxations and/or self-labeling) } \\\hline
     ScribbleSup \cite{lin2016scribblesup} & V2 (VGG16) & 8 & - & \checkmark & - & DN & 63.1 \\
     DenseCRF loss$^*$  \cite{tang2018regularized} & V3+ & 12 & \checkmark & - & - & DN & 75.8 \\
     GridCRF loss$^*$ \cite{marin2021robust} & V3+ & 12 & - & \checkmark & - & NN & 75.6 \\
     NonlocalCRF loss$^*$ \cite{veksler2022sparse} & V3+ & 12 & \checkmark & - & - & SN & 75.7\\
     \rowcolor{gray!15} $\mathbf{H_\CCE+P_\Q}$ & V3+ & 12 & - & - & \checkmark & NN & \textcolor{teal}{77.5} \\
     \rowcolor{gray!15} $\mathbf{H_\CCE+P_\CD}$ & V3+ & 12 & - & - & \checkmark & NN & \textcolor{teal}{77.7} \\
     \rowcolor{gray!15} $\mathbf{H_\CCE+P_\CD}$ (no pretrain) & V3+ & 12 & - & - & \checkmark & NN & \textcolor{teal}{76.7} \\
     \rowcolor{gray!15} $\mathbf{H_\CCE+P_\CD}$ (no pretrain) & V3+ & 16 & - & - & \checkmark & NN & 77.6 \\
     \rowcolor{gray!15} $\mathbf{H_\CCE+P_\CD}$ & V3+ & 16 & - & - & \checkmark & NN & 78.1 \\
     \rowcolor{gray!15} $\mathbf{H_\CCE+P_\CD}$ (no pretrain) & V3+ & 16 & - & - & \checkmark & NN & 77.6 \\
     \rowcolor{gray!15} $\mathbf{H_\CCE+P_\CD}$ (no pretrain) & ViT-linear & 12 & - & - & \checkmark & NN & 80.8 \\
     \rowcolor{gray!15} $\mathbf{H_\CCE+P_\CD}$ (no pretrain) & ViT-linear & 16 & - & - & \checkmark & NN & 80.94 \\\hline
    \end{tabular}}
    \caption{Comparison of scribble-supervised segmentation methods (without CRF postprocessing). We focus on general solutions applicable to any standard architecture based on mathematically well-founded convergent training procedures optimizing a clearly defined loss (the last block). The second block shows specialized solutions based on architectural modifications and/or ad-hoc training procedures that are often non-convergent and hard to reproduce. The numbers are mIoU on the validation dataset of Pascal VOC 2012. The backbone is ResNet101 unless stated otherwise. V2: deeplabV2. V3+: deeplabV3+. ViT-linear: ViT backbone and linear decoder. $\mathcal{N}$: neighborhood. ``$*$'': reproduced results. GD: gradient descent. SL: self-labeling. ``no pretrain'' means the segmentation network is not pretrained using cross-entropy on scribbles. In some cases (V3+, batch size 12) soft self-labeling outperforms full supervision (see teal color numbers).}
    \label{tab: SOTA} \vspace{-1ex}
\end{table*}

\subsection{Soft vs. hard self-labeling vs. gradient descent} \label{sec: soft vs hard vs GD}

Here we review our general observations about the Potts-based WSSS using the scribble-based results summarized 
in Table \ref{tab: soft sl vs hard sl vs gd}. First, in the simplest approach 
using direct optimization of the network parameters w.r.t. Potts-regularized losses like \eqref{eq:loss_ws} 
and stochastic gradient descent, one benefits from a larger neighborhood size \cite{tang2018regularized}.
They use $P_\BL$ where DN smoothens the Potts model \cite{marin2019beyond} helping SGD to avoid local minima.
Higher-order optimization based on auxiliary self-labeling loss \eqref{eq:loss_self} significantly benefits from soft pseudo-labels since they can represent uncertainty. In this case (normalized) quadratic Potts relaxations offer two advantages over the tight bilinear relaxation that leads to hard solutions. On the one hand, 
they do not suppress label softness; this softness is even formally related to classification uncertainty in the random walker method \cite{grady2006random}. On the other hand, the convexity of quadratic relaxation does not require DN to simplify loss for gradient descent. Then, it is not surprising that NN works better than DN. 
The latter makes the Potts model behave as a cardinality potential \cite{veksler2019efficient}, 
while in the case of NN the Potts model directly regularized the geometric boundary of objects \cite{boykov2003geodesic}.
Figure~\ref{fig: soft sl vs hard sl vs gd} also compares these approaches across different scribble lengths.

\subsection{Comparison to SOTA}
\label{sec: experiment SOTA}
This section uses DeepLabV3+ with ResNet101 backbone and ViT-linear transformer to evaluate our self-labeling methodology on these standard architectures and to compare with representative SOTA for scribble-based segmentation. 
Table \ref{tab: SOTA} presents only the results before any post-processing. Our method may outperform the fully-supervised method (teal color numbers for DeepLabV3+ and batch size 12). Our ``text-book'' training technique, i.e. minimization of a well-defined loss, applied to standard architectures outperforms complex WSSS systems requiring architectural modifications and/or ad-hoc multi-stage training procedures.

\section{Conclusions}
\label{sec: conclusion}
This paper proposes a mathematically well-founded self-labeling WSSS framework for standard segmentation networks based on a joint loss \eqref{eq:loss_self} w.r.t. network predictions and soft {\em pseudo-labels}. The latter was motivated as auxiliary variables simplifying optimization of \eqref{eq:loss_ws}.
Conceptual properties and systematic evaluation of the terms in our soft self-labeling loss \eqref{eq:loss_self} advocate for the {\em collision cross-entropy} and {\em collision divergence}. The former replaces the standard cross-entropy $H$, and the latter is used as a log of a normalized quadratic relaxation of the Potts model $P$ over the NN neighborhood. Our general methodology works with any standard architecture and outperforms complex specialized systems based on architectural modifications and/or ad-hoc multi-stage non-convergent training. In contrast, our ideas are easy to understand, reproduce, and extend to other forms of weakly-supervised segmentation (boxes, class tags, etc.).

\clearpage
{
    \small
    \bibliographystyle{main}
    \bibliography{main}
}

\clearpage
\appendix
\section{Optimization Algorithm} 
\label{sec: optimization algorithm}
In this section, we will focus on the optimization of our loss where we iterate the optimization of $y$ and $\sigma$. 
The network parameters are optimized by standard stochastic gradient descent in all our experiments. Pseudo-labels are also estimated online using a mini-batch.
To solve $y$ at given $\sigma$, it is a large-scale constrained convex problem. While there are existing general solvers to find global optima, such as projected gradient descent, it is often too slow for practical usage. Instead, we reformulate our problem to avoid the simplex constraints so that we can use standard gradient descent in PyTorch library accelerated by GPU. Specifically, instead of directly optimizing $y$, we optimize a set of new variables \{$l_i\in \mathbb{R}^K$, $i\in\Omega$\} where $y_i$ is computed by $softmax(l_i)$. Now, the simplex constraint on $y$ will be automatically satisfied. Note that the hard constraints on scribble regions still need to be considered because the interaction with unlabeled regions through pairwise terms will influence the optimization process. Inspired by \cite{zhu2002learning}, we can reset $softmax(l_i)$ where $i\in S$ back to the ground truth at the beginning of each step of the gradient descent.   

However, the original convex problem now becomes non-convex due to the Softmax operation. Thus, initialization is important to help find better local minima or even the global optima. Empirically, we observed that the network output logit can be a fairly good initialization. The quantitative comparison uses a special quadratic formulation where closed-form solution and efficient solver \cite{grady2005multilabel,grady2006random} exist. We compute the standard soft Jaccard index for the pseudo-labels between the solutions given by our solver and the global optima. The soft Jaccard index is 99.2\% on average over 100 images. In all experiments, the number of gradient descent steps for solving $y$ is 200 and the corresponding learning rate is 0.075. To test the robustness of the number of steps here, we decreased 200 to 100 and the mIoU on the validation set just dropped from 71.05 by 0.72. This indicates that we can significantly accelerate the training without much sacrifice of accuracy. When using 200 steps, the total time for the training will be about 3 times longer than the SGD with dense Potts \cite{tang2018regularized}.

\section{Additional Experiments}
\label{sec: experimental settings}
\paragraph{Dataset and evaluation}
We mainly use the standard PASCAL VOC 2012 dataset \cite{everingham2009pascal} and scribble-based annotations for supervision \cite{lin2016scribblesup}. The dataset contains 21 classes including background. Following the common practice \cite{chen2014semantic,tang2018normalized,tang2018regularized}, we use the augmented version which has 10,582 training images and 1449 images for validation. We employ the standard mean Intersection-over-Union (mIoU) on validation set as the evaluation metric.
We also test our method on two additional datasets. One is Cityscapes \cite{Cordts2016Cityscapes} which is built for urban scenes and consists of 2975 and 500 fine-labeled images for training and validation. There are 19 out of 30 annotated classes for semantic segmentation.
 The other one is ADE20k \cite{zhou2017scene} which has 150 fine-grained classes. There are 20210 and 2000, images for training and validation. Instead of scribble-based supervision, we followed \cite{liang2022tree} to use the block-wise annotation as a form of weak supervision.

\begin{table}[h]
    \centering
    \resizebox{0.5\textwidth}{!}{\begin{tabular}{c|c|c|c}\hline
       Method  &  Architecture & Cityscapes & ADE20k \\\hline
       \multicolumn{4}{c}{\bf Full supervision} \\\hline
       Deeplab \cite{chen2018encoder} & V3+ & 80.2 & 44.6 \\\hline
       \multicolumn{4}{c}{\bf \
       Block-scribble supervision} \\\hline
       DenseCRF loss \cite{tang2018regularized} & V3+ & 69.3 & 37.4 \\
       GridCRF loss$^*$ \cite{marin2021robust} & V3+ & 69.5 & 37.7 \\
       TEL \cite{liang2022tree} & V3+ & 71.5 & 39.2 \\
       \rowcolor{gray!15} $\mathbf{H_\CCE+P_\CD}$ & V3+ & 72.4 & 39.7 \\\hline
    \end{tabular}}
    \caption{Comparison to SOTA methods (without CRF postprocessing) on segmentation with block-scribble supervision. The numbers are mIoU on the validation dataset of cityscapes \cite{Cordts2016Cityscapes} and ADE20k \cite{zhou2017scene} and use $50\%$ of full annotations for supervision following \cite{liang2022tree}. The backbone is ResNet101. ``$*$'': reproduced results. All methods are trained in a single-stage fashion.}
    \label{table: segmentation different datasets}
\end{table}


\end{document}